%% file: main.tex
\documentclass[a4paper, 10 pt, conference]{ieeeconf}

\IEEEoverridecommandlockouts                            
\usepackage[utf8]{inputenc} 
\usepackage[T1]{fontenc}    
\usepackage{hyperref}       
\usepackage{url}            
\usepackage{booktabs}       
\usepackage{amsfonts}       
\usepackage{nicefrac}       
\usepackage{microtype}      
\usepackage{lipsum}
\usepackage[T1]{fontenc}

\usepackage{cite}

\usepackage{amsmath,amssymb}
\usepackage{algorithmic}
\usepackage{graphicx}
\usepackage[table,xcdraw]{xcolor}
\usepackage{subfigure}
\usepackage{textcomp}
\usepackage{gensymb}
\usepackage{epsfig} 
\usepackage{xcolor}

\usepackage{booktabs}
\usepackage{array}
\usepackage{multirow}

\usepackage{ulem}
\usepackage{units}
\newcommand{\subparagraph}{}

\title{\LARGE \bf
The Jamming Donut: A Free-Space Gripper based on Granular Jamming.
}

\author{Therese Joseph$^{1,2}$, Sarah Baldwin$^{1}$, Lillian Guan$^{1,3}$, James Brett$^{1}$ and David Howard$^{1}$
\thanks{$^{1}$ CSIRO, Australia; contact {david.howard@csiro.au}}
\thanks{$^{2}$ Queensland University of Technology, Australia}
\thanks{$^{3}$ Griffith University, Australia}
}

\begin{document}

\maketitle
\thispagestyle{empty}
\pagestyle{empty}


\input{Content/1.0_Abstract}

\input{Content/2.0_Introduction}

\input{Content/3.0_Background}

\input{Content/4.0_Methodology}

\input{Content/5.0_Results}

\input{Content/6.0_Discussion}

\addtolength{\textheight}{-12cm}
\end{document}

%% file: Content/1.0_Abstract.tex
\begin{abstract}
 Fruit harvesting has recently experienced a shift towards soft grippers that possess compliance, adaptability, and delicacy. In this context, pneumatic grippers are popular, due to provision of high deformability and compliance, however they typically possess limited grip strength.  Jamming possesses strong grip capability, however has limited deformability and often requires the object to be pushed onto a surface to attain a grip.  This paper describes a hybrid gripper combining pneumatics (for deformation) and jamming (for grip strength).  Our gripper utilises a torus (donut) structure with two chambers controlled by pneumatic and vacuum pressure respectively, to conform around a target object. The gripper displays good adaptability, exploiting pneumatics to mould to the shape of the target object where jamming can be successfully harnessed to grip. The main contribution of the paper is  design, fabrication, and characterisation of the first hybrid gripper that can use granular jamming in free space, achieving significantly larger retention forces compared to pure pneumatics.  We test our gripper on a range of different sizes and shapes, as well as picking a broad range of real fruit.

\end{abstract}


%% file: Content/2.0_Introduction.tex
\section{INTRODUCTION}

\begin{figure}[h]
\label{fig:fig1}
    \centering
    \includegraphics[width=0.7\columnwidth]{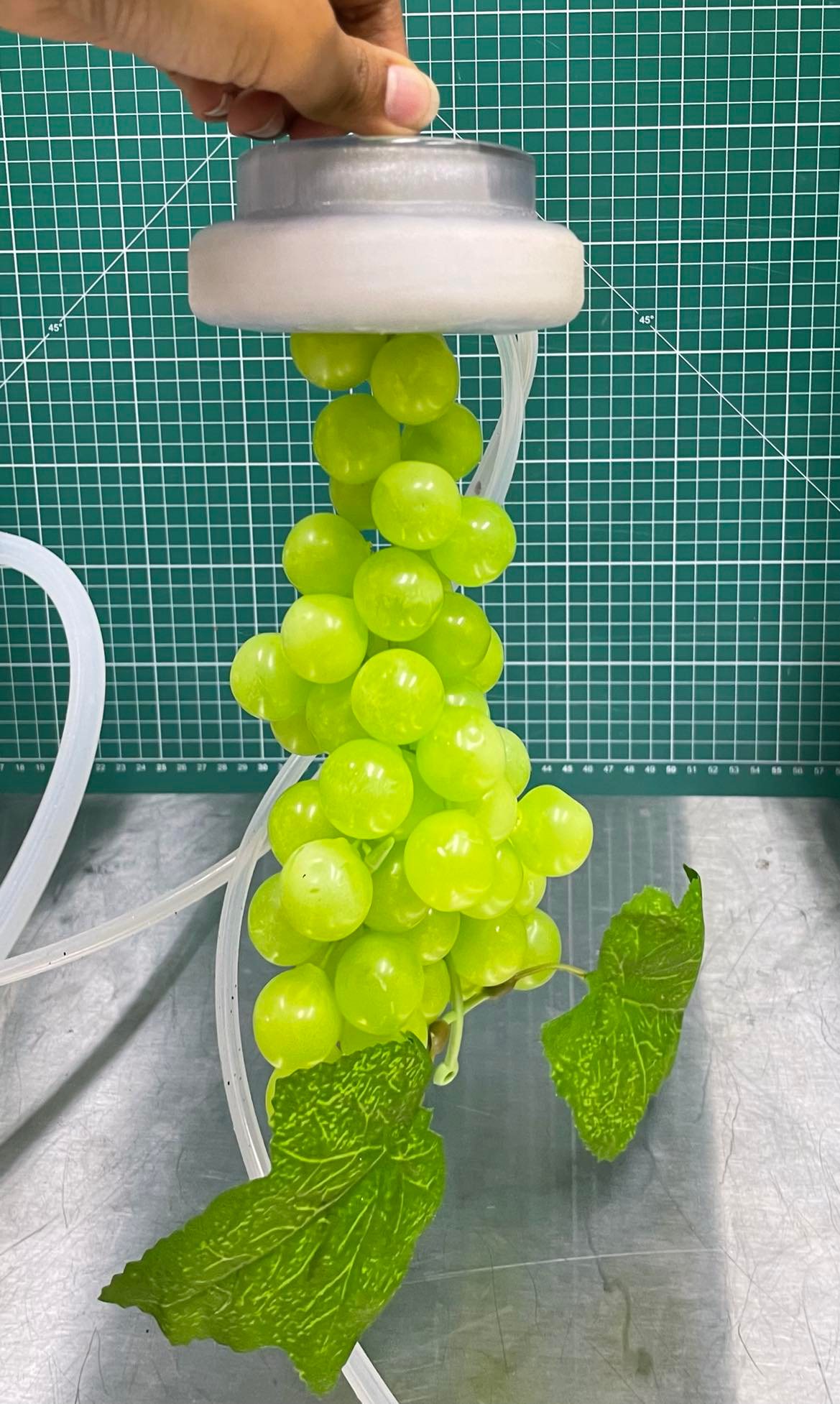}
    \caption{Demonstrating our jamming donut gripper.}
    \label{fig:retentionsetup}
\end{figure}

Soft grippers are poised to revolutionise a number of industries where compliance and robustness are key \cite{shintake2018soft}.  One such industry is agriculture, where crops of soft produce (e.g., fruit) must be picked without bruising, and where the target object may display significant variation of size and shape, even within a single species.  Development of suitable soft systems for these real world tasks remains an open challenge, but typically such systems must strike a balance between deformation (the ability to adapt to shape and size variations in the target objects) and variable stiffness \cite{manti2016stiffening}, which imparts the ability to grasp the object with sufficient strength to perform followup tasks as part of the harvesting process, such as twisting and pulling \cite{navas2021soft}).  

Standard soft gripping techniques lack this ability when used in isolation.  Pneumatic actuation \cite{xavier2022soft} provides high deformability and compliance, but lacks the ability to exert large grip forces.  Conversely, stiffness variation methods such as granular jamming are capable of providing high grip forces, but typically provide low deformation (and typically low negative deformation, e.g., loss of volume, of $\approx 3\%$ \cite{brown2010universal}) when a jamming transition is used to achieve the necessary grip forces.  This type of granular jamming, e.g., that seen in the universal gripper \cite{brown2010universal}, also requires the target object to be sandwiched against a surface so that the gripper can push into the object, which makes it difficult to harness the benefits of its excellent stiffness variation in open space scenarios.  

Here we present a hybrid soft gripper that achieves the deformation/stiffness balance through a combination of pneumatics (for deformation) and jamming \cite{fitzgerald2020review} (for stiffness variation and grip strength).  The key functional principle of our gripper (Fig.~\ref{fig:fig1}) -- the Jamming Donut -- is the ability to inflate the outer chamber of a ring-style gripper against a rigid case, which forces an inner chamber filled with granular material to constrict against the target object.  Subsequently applying vacuum pressure to the inner chamber jams the granular material whilst retaining pressure against the target object's surface.  This results in a gripper that can adapt to a variety of different sizes and a range of shapes, whilst simultaneously allowing for large grip forces between the target object and the jammed inner chamber.  Importantly, our gripper is capable of free-space gripping, and thus provides a route towards harnessing jamming for gripping and harvesting soft produce.  The key contributions of this paper are:

\begin{itemize}
    \item Design of the first torus free-space gripper using granular jamming.
    \item Development of a novel grip strategy based on joint inflation and jamming, with high observed performance.
    \item Extensive experimentation on performance relative to target object shape and size, with characterisation of pressure exerted on the object through custom sensorised test objects.
\end{itemize}

In the remainder of the paper we discuss relevant literature (Section II).  We then describe the design and fabrication of the gripper (Section III), and report on experiments showing successful grasping of a range of object shapes and sizes, with particular attention paid to analysis of grip strength and gripper deformation.  We also show the capability to grasp a range of different fruit (Section IV).  We then discuss the results, and provide some future outlook (Section V).

%% file: Content/3.0_Background.tex
\section{BACKGROUND}

\subsection{Hybrid Jamming}

Hybrid mechanisms are increasingly popular in the literature as a way to achieve desired behaviours with appropriate levels of deformation and stiffness by combining multiple mechanisms, e.g., pneumatics with cable-driven actuation \cite{kim2020soft} to guide inflation-induced deformations along the desired trajectory.

Jamming is popular as a technique to provide variable stiffness when combined with actuation mechanisms.  Amongst the variable stiffness mechanisms, it is relatively fast to transition from minimally to maximally stiff (2-3s), and displays large stiffness variation between these two extremes.  It is also simple to integrate with various actuation mechanisms due to its relatively unconstrained design possibilities, offering a simple route to hybridisation \cite{fitzgerald2020review}.  

Modular vacuum/jamming systems are shown to achieve target deflections and subsequently jam to hold their shape \cite{Robertsoneaan6357}.  Hybrid manipulation is a popular field, with examples of actuation combined with jamming \cite{wei2016novel}, and hydraulic deflection with a central jamming stiffening channel \cite{cianchetti2013stiff} shown to provide high dexterity for surgical operations.  Jamming has also been combined with pneumatic muscles \cite{al2018variable}, and with suction \cite{suction_gripper}.  

The literature also shows examples of a snake-like manipulator is presented that combines layer jamming with wire actuation \cite{tubular_snake}.  Numerous works on bio-inspired robotic fingers have combined, e.g., two jamming mechanisms \cite{yang2019hybrid}, jamming and tendon actuation \cite{mizushima2018multi}, and jamming with pneumatics \cite{li2018distributed}.

The choice of actuation mechanism to pair with the jamming component comes down to the desired range and type of motion, (e.g., linear, compressive, bending) \cite{xavier2022soft}.  Given our choice of a torus-style gripper targeting a range of object shapes and sizes, compressive actuation is required, indicating the use of pneumatics.

\subsection{Soft robotics for agriculture}

Robotic agriculture is a rapidly-growing field, promising efficiency and scale \cite{zhou2022intelligent}, which have been deployed across a range of target crops including kiwi fruit \cite{kiwi_fruit} and tomatoes \cite{tomatoes}.  As well as gripping and picking, solutions may include, e.g.  cutting \cite{birrell2020field}.

The largest challenge for a gripping end effector is the variation in shape and size, even among fruits belonging to the same type.   Furthermore, the viscoelastic property of fruits means that the gripper contact point and applied pressure needs to be considered to avoid bruising.  As such, soft gripping has a prominent place within this field, offering rapid, robust gripping without the risk of damaging soft produce \cite{navas2021soft}.  

Successful applications of soft gripping to handling soft produce include the development of a diaphragm gripper with a reconfigurable modular architecture to surround the object  \cite{navas2021diaphragm}, as well as a pneumatic 4-chambered circular gripper that inflates to narrow a central aperture and cause gripping to occur \cite{wang2021circular}.  The benefit for soft produce handling is that the necessary grip force can be spread across multiple contact points and a larger surface area, reducing the risk of bruising.  The latter is the most similar to our approach \cite{wang2021circular}, being an inflation-driven torus gripper.  However, this approach is purely pneumatic and contains no stiffening element.  Our differentiator is hybridisation with a granular jamming mechanism, and our results show the benefits of this novelty through the potential to generate significantly increased grip forces.

Overall, we note that (i) circular or torus grippers are popular in the literature for handling soft produce, with promising results reported. (ii) hybridisation using jamming is a compelling avenue to realise a variable-stiffness gripper, that can be combined with pneumatics to provide both deformation and heightened grip strength. Specifically when integrated into a torus gripper, it enables jamming to occur in free space, as is required for numerous agricultural gripping tasks. Our original contribution builds on the state of the art by hybridising a pneumatic torus gripper with jamming.

%% file: Content/4.0_Methodology.tex
\section {Materials and Methods}

\subsection{Manufacture}

\begin{figure}[t!]
    \centering
    \includegraphics[width=\linewidth]{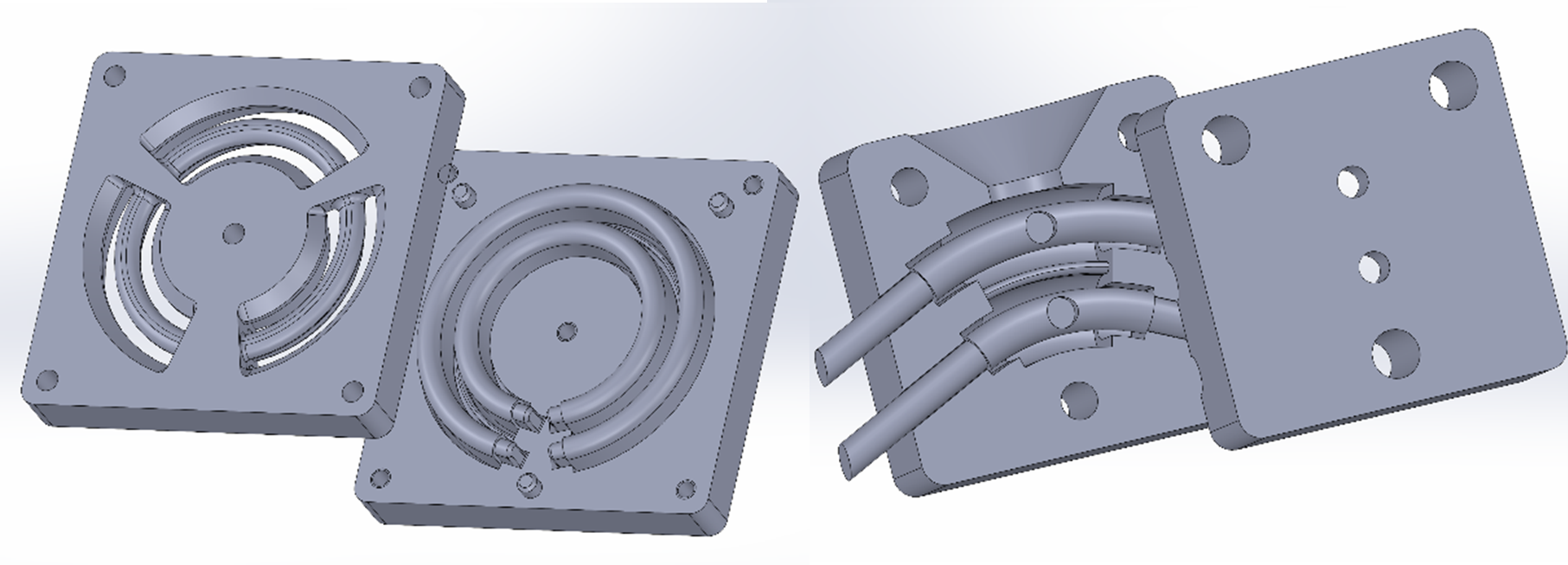}
    \caption{Four-part moulds: main (left), joiner (right).}
    \label{fig:moulds}
\end{figure}

\begin{figure}[t!]
    \centering
    \includegraphics[width=0.7\linewidth]{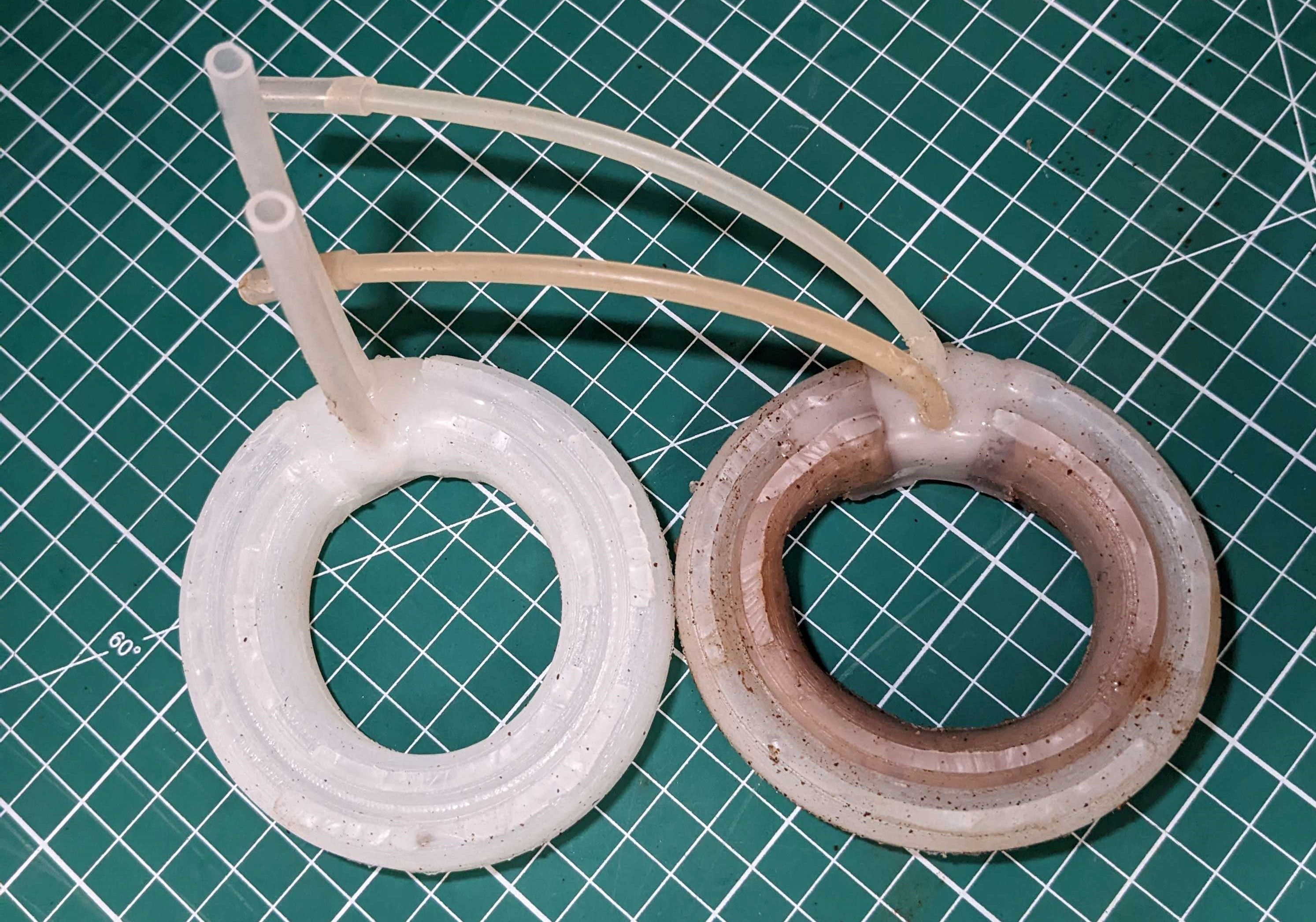}
    \caption{Cast grippers: empty (left) and with the inner chamber filled with ground coffee (right).}
    \label{fig:donuts}
\end{figure}

Our donut gripper is manufactured in cast silicone using  two 3D printed moulds (Fig.~\ref{fig:moulds}); the main four part dual chamber donut mould, and a separate four part joining mould to seal the gripper and attach the pressure control tubes. The moulds feature multiple alignment tabs and tight tolerances to ensure accurate casting (Fig.~\ref{fig:donuts}).

\begin{figure}[t!]
    \centering
    \includegraphics[width=0.7\linewidth]{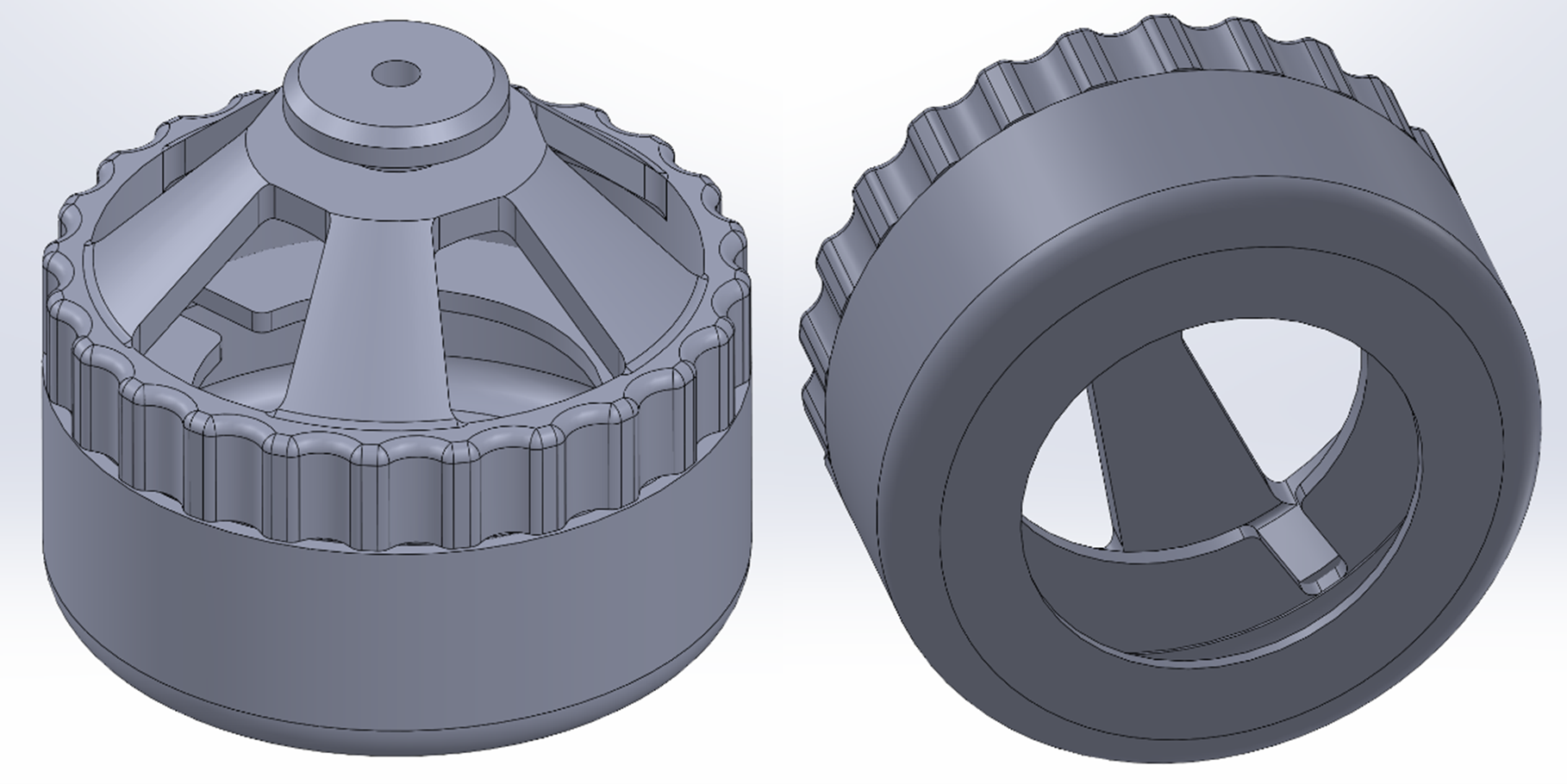}
    \caption{CAD render of the rigid strain limiting shell.}
    \label{fig:shell}
\end{figure}

The gripper is encased in a rigid 3D printed outer shell that acts as a strain limiting layer and provides a means of attaching to the test rig (Fig.~\ref{fig:shell}). The shell fully encompasses the silicone chambers to confine any pneumatic action to produce inward expansion. The shell and moulds are printed in ASA on a Stratasys F370 printer.

The silicone chambers are cast in Smooth-On Ecoflex 00-30, which is treated in a vacuum-based degassing chamber for 5 minutes before being poured into the moulds and left to set for 4 hours.  After removal from the moulds, the inner chamber is filled with 8g of coffee grounds using a funnel.  The Chambers are sealed by attaching the silicone pressure lines and end cap.  Finally, the cast gripper is placed in the rigid shell.

\subsection{Test Setup}
\label{sec:TS}
\begin{figure}[h]
    \centering
    \includegraphics[width=0.8\linewidth]{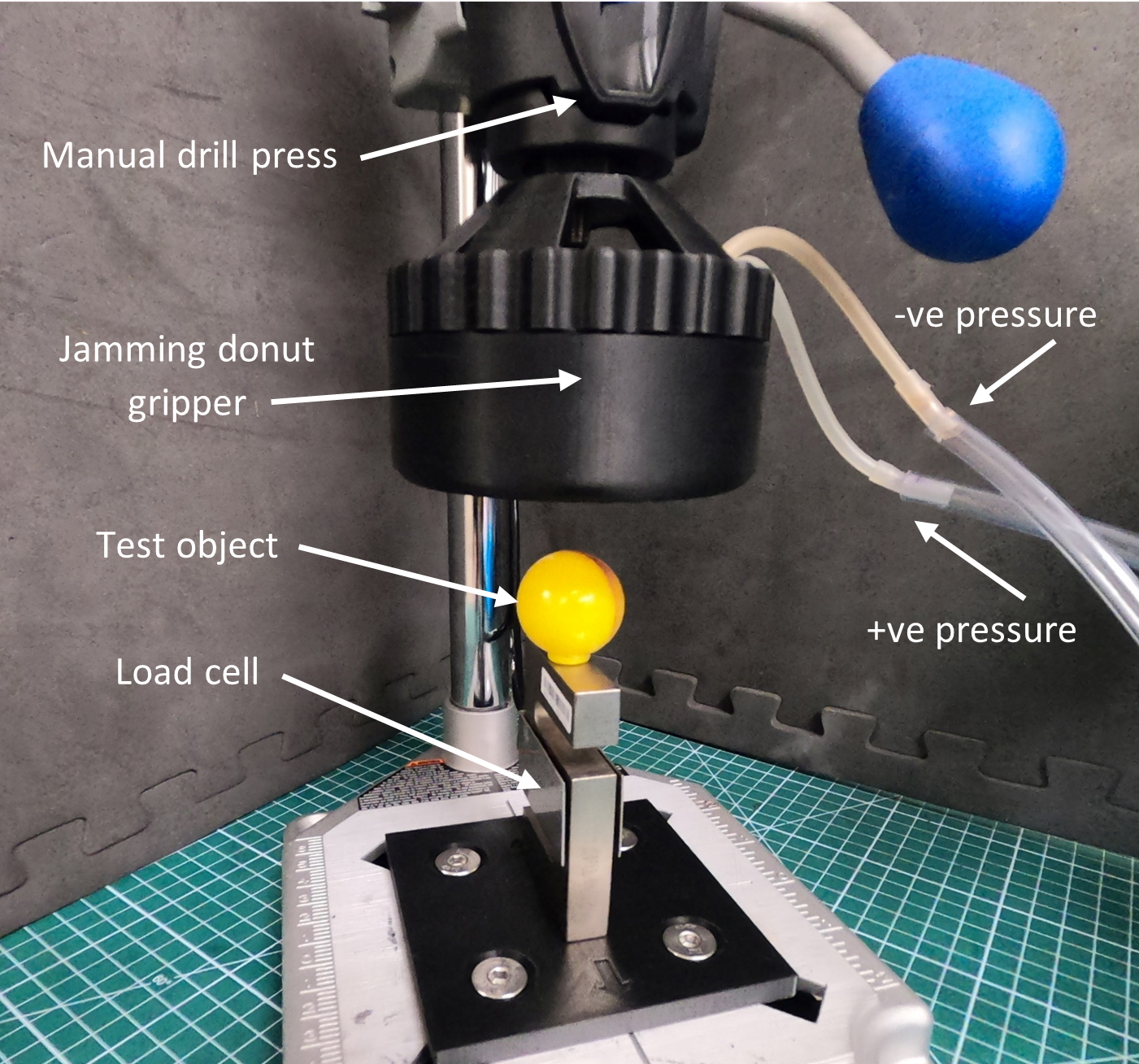}
    \caption{Test Rig: the silicon chambers are inserted into the rigid shell, which attaches to the drill press.  Positive and negative pressure lines are attached.  A test object is screwed into the load cell via a 3D printed thread.}
    \label{fig:retentionsetup}
\end{figure}

We perform a range of tests to characterise our gripper, using a common testing setup as depicted in Fig.~\ref{fig:retentionsetup}. In this setup, the gripper is bolted to a Dremel drill press stand via the rigid shell.  A positive pressure line is connected to the outer chamber, and a negative pressure line with in-line filter to the inner chamber.  The other end of the negative pressure line is attached to a Rocker 300 vacuum pump.  

The selected test object is attached to a Zemic H3-C3-25kg-3B load cell via 3D printed thread. The load cell is secured to the drill press stand via a metal mounting plate.

During data collection, the pressure control of the gripper is automated using an Arduino controlled pressure regulation and switching board, ensuring a consistent testing procedure by activating the pressure lines as required.  

To perform a test, the gripper is lowered over the test object.  The outer chamber is then inflated for 5s and held while the inner negative chamber is activated with pressure of -80kPa.  The lever is then slowly released, allowing the gripper to be pulled off the test object.  The retention force data was recorded from the load cell using a Raspberry Pi 3 B+ via a Sparkfun load cell amplifier.  At the conclusion of each test, both chambers were reset to atmospheric pressure.

%% file: Content/5.0_Results.tex
\section{Experimentation and Results}

\subsection{Comparison of Jamming Methods}
The ability to generate grip force across a range of object sizes is a key requirement of our gripper. We use a set of 3D printed spherical test objects in a range of diameters (20, 25, 30, 35, 40mm).  Spheres are chosen as they are representative of the shape of many fruits.  To provide comparative baseline performance, we compare a number of gripper configurations. 

\begin{itemize}
    \item Active jamming, as detailed in Sec~\ref{sec:TS}
    \item Passive jamming, only applying positive pressure, allowing the inflation of the outer chamber to passively jam the granular material against the object.
    \item No jamming. Both chambers are empty, and positive pressure is applied to the outer chamber only.  This is purely pneumatic actuation, following \cite{wang2021circular}.\footnote{Preliminary testing investigated inflating both chambers, however results were worse due to beneficial effects of a compliant contact created between the uninflated inner chamber and  the test object}.
\end{itemize}

\begin{figure}[ht]
    \centering
    \includegraphics[width=1\linewidth]{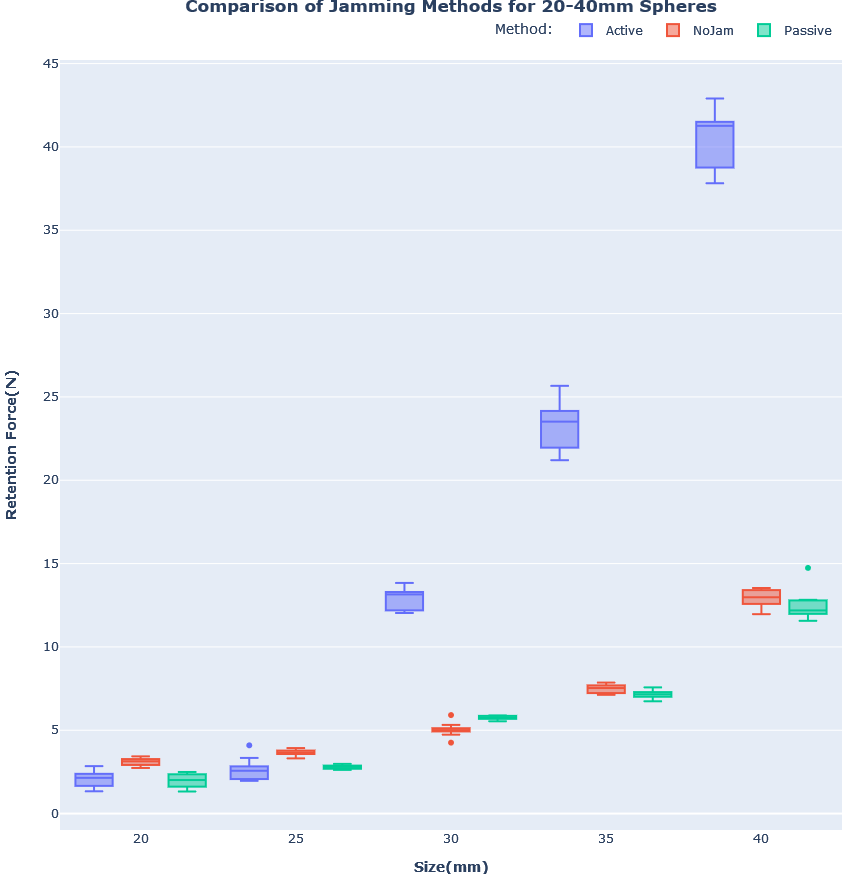}  
    \caption{Boxplot showing increasing grip forces (N) with increasing object size amongst the three gripper configurations tested.  Active jamming is seen to have a large impact on grip strength on objects 30mm and over.  Bars denote standard error.}
    \label{fig:retentionresults}
\end{figure}

Each combination of gripper configuration and object size was repeated 10 times, with the results averaged.  Overall, we see that all configurations can generate grips on all test objects, e.g., contact is made between the inner chamber and test object in all cases -- see Fig.~\ref{fig:retentionresults}.  Grip forces above 10N are not observed for 20mm and 25mm objects; this relates to the maximum deformation range achievable using our inflation-based actuation mechanism.  For these objects, all three configurations provide similar grip strength, with a slight preference in terms of performance and standard error to the purely pneumatic {\it No jamming} configuration, as the addition of coffee in the inner chamber limits the inwards deformation of the chambers when compressing onto the object.  This effect is negligible in terms of performance.

Passive jamming is seen to have be broadly comparable to no jamming throughout the object sizes considered.  

Active jamming generates large grip forces on any object over 30mm, far in excess of the other two configurations.  Active jamming results exponentially increase with size of test object, compared to linear increases of passive jamming and no jamming methods as size increases.  

On 30mm objects, mean grip forces are $\approx$13N, compared to $\approx$5N and $\approx$6N for no jamming and passive jamming respectively.  The effects become more pronounces as object size increases, as it is easier to create large surface area contacts between gripper and object to more readily exploit the jamming mechanism. This is reflected in 35mm objects ($\approx$24N compared to $\approx$7N and $\approx$7N) and 40mm objects ($\approx$41N compared to $\approx$13N and $\approx$12N), for active, no jamming, and passive jamming respectively.

\subsection{Generalised grasping}
\begin{figure}[t]
    \centering
\subfigure[]{\includegraphics[width=1\linewidth]{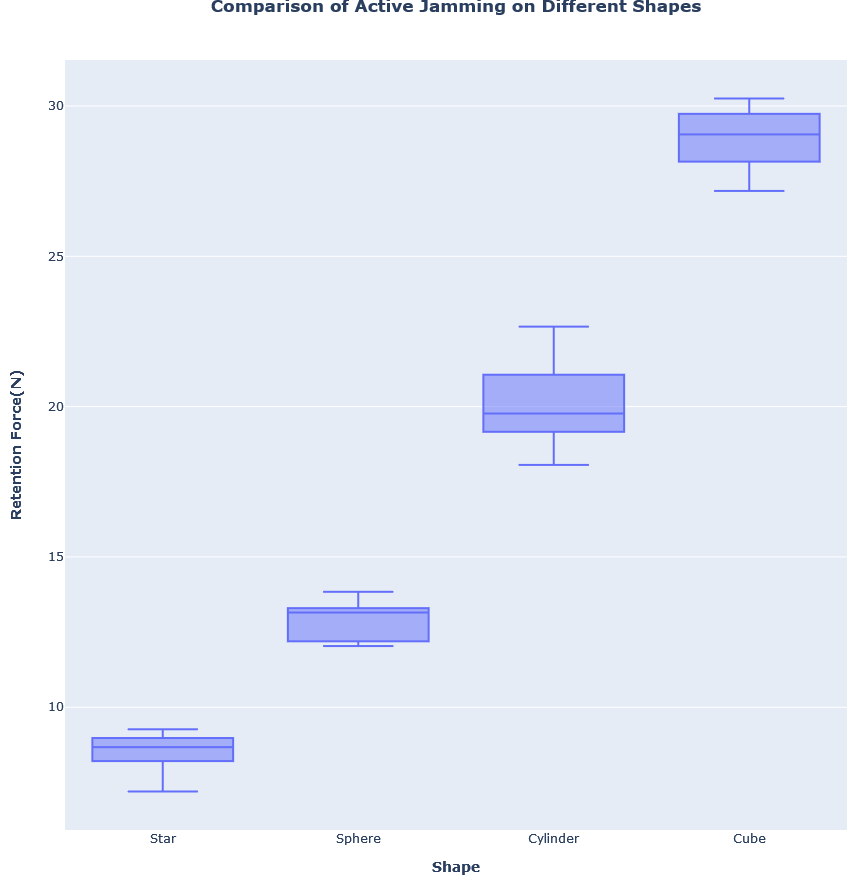} }\\
\subfigure[]{\includegraphics[width=.9\linewidth]{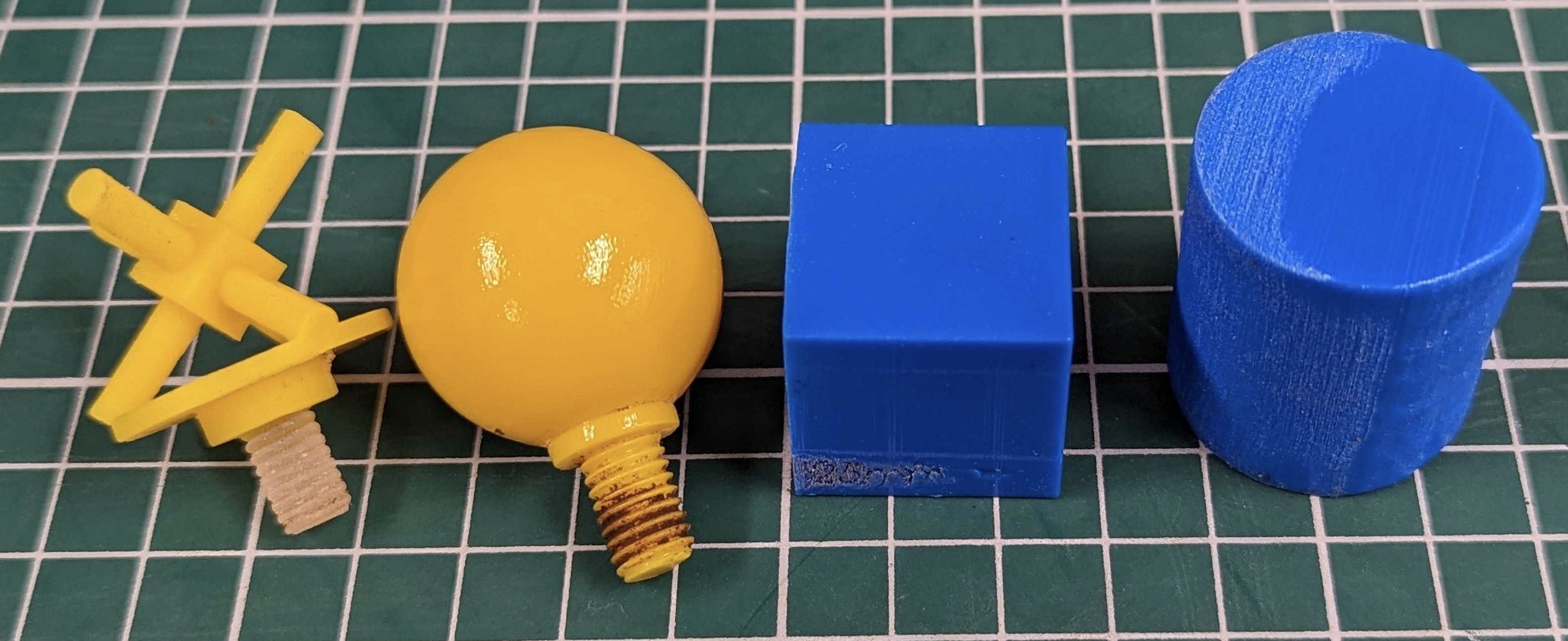} }
    \caption{(a) Boxplot showing grip strength attained on a range of 30mm test objects by the active jamming configuration of the gripper.  Mean grip forces range between 8.6N (star) and 29.0N (cube). Bars denote standard error. (b) The test objects selected to present a diverse set of challenges to the gripper.}
    \label{fig:shapegrips}
\end{figure}

The generalisability of the gripper to differently-shaped objects is assessed (Fig.~\ref{fig:shapegrips}(a)) using 30mm diameter star, sphere, cylinder and cube shapes (Fig.~\ref{fig:shapegrips}(b)). This size was selected as the first size at which jamming effects become prominent, so any sensitivity to object geometry is likely to be observable at this size.  Each test was repeated 10 times and the results averaged.%

Fig.~\ref{fig:shapegrips}(a) shows a pattern of performance, with the strongest grip on the cube (29.0N) followed by the cylinder (19.7N), the sphere (13.1N) and the star (8.6N). The cube has the most surface area contact, and generates strong grips by (partially) enveloping the corners.  The cylinder performs better than the sphere due to presenting more easily accessible surface area to the gripper. Our gripper struggles to expand into the small nooks of the star shape.  Overall error is relatively low, with over 50\% of the data falling within less than 2N of the average.

\subsection{Grip comparison to universal jamming gripper}
To assess the free space functionality of our novel jamming gripper, we compare representative force exertion profiles generated during a grip to that of a standard universal jamming gripper (e.g., \cite{brown2010universal}).  The comparative gripper was created by filling a 12.5cm latex balloon with ground coffee, and attaching to the drill press stand via a 3D printed adapter.  This size of gripper is considered appropriate for the objects being gripped \cite{brown2010universal}.

\begin{figure}[h]
    \centering
    \includegraphics[width=1\linewidth]{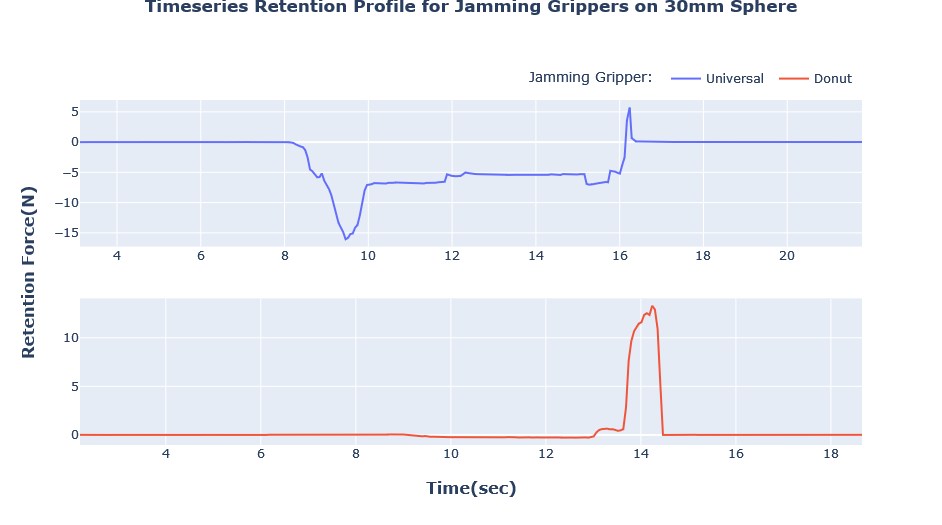}  
    \caption{Comparing typical force exertion profiles generated during a grip to that of an equivalent universal jamming gripper.  Our gripper displays minimal activation force, which is the dominant force in the universal gripper, and stronger grip force (14.2N compared to 5.2N).}
    \label{fig:timeseriesretentionprofile}
\end{figure}

The time series profile illustrated in Fig.~\ref{fig:timeseriesretentionprofile} shows the significantly different force profiles for the universal jamming gripper compared to the jamming donut gripper.  The universal jamming gripper shows a well-studied force profile, generating 16.1N of activation force as the gripper is pushed against the object and deforms around it (8s to 9.5s).  The vacuum pump is then activated and the gripper slowly raised until the grip is broken (from 16s), generating a maximum grip force of 5.2N.

Note that the activation force is typically something to be minimised; it is an artefact of using universal jamming grippers and techniques are frequently used to minimise this force, e.g, fluidisation via positive pressure \cite{amend2012positive} or vibration \cite{mishra2021vibration}.  Our gripper generates no noticeable activation force, and additionally generates a grip force from 13s to 14.5s of 14.2N - nearly three times of the comparative gripper, but still less than the 'cost' of activation force in the universal gripper.  This is important for soft produce handling; no superfluous forces are generated during a grip, which would practically reduce the chance of bruising.  We also note the large contact surface area presented by our gripper compared to the universal gripper, meaning the forces have to opportunity to be more spread across the contacted area of the test object.

\subsection{Pressure profiles}

\begin{figure}[h]
    \centering
    \includegraphics[width=0.6\linewidth]{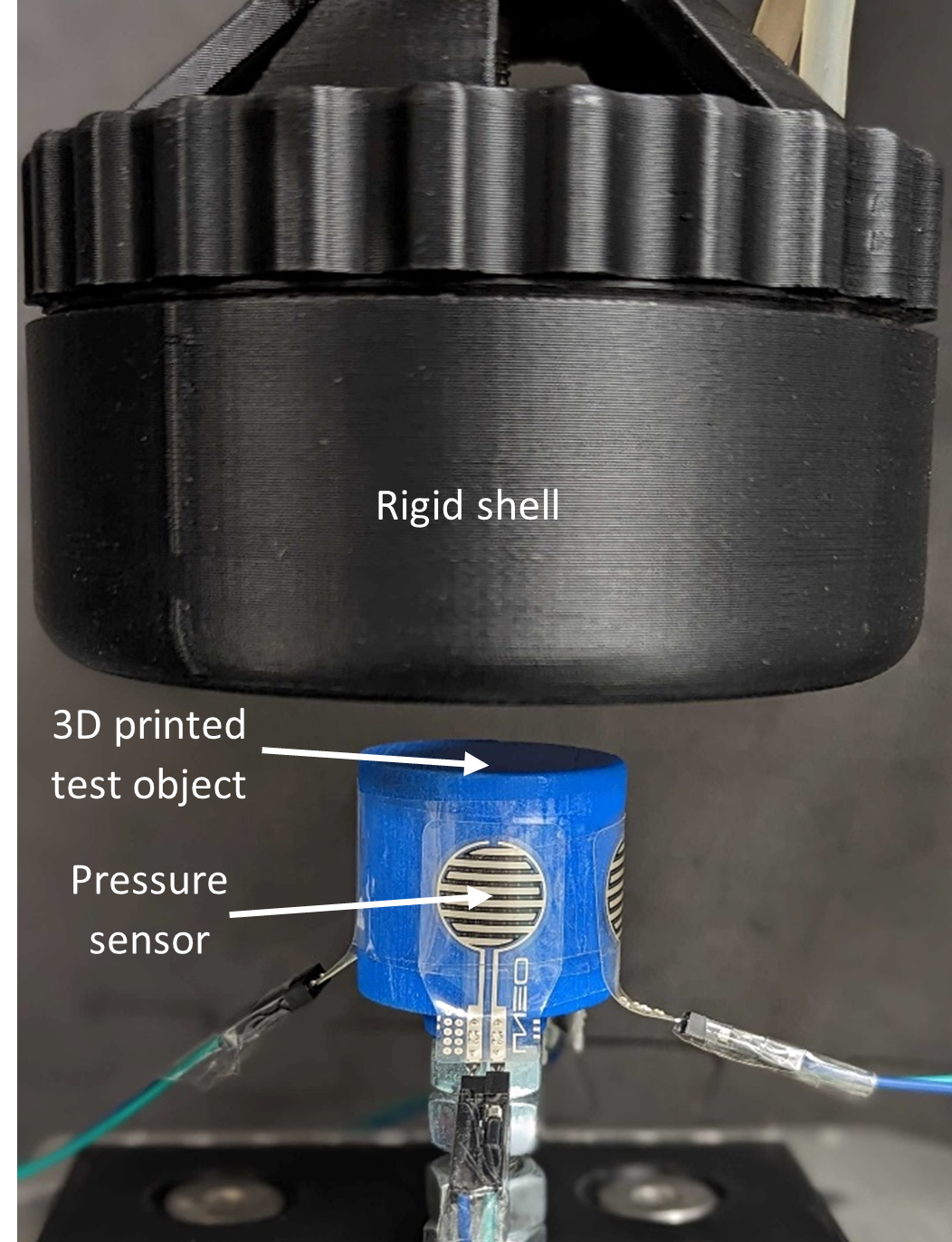}  
    \caption{Pressure Test Rig, showing the placement of the pressure sensors on the 3D printed cylinder.}
    \label{fig:pressuresetup}
\end{figure}

To complete our characterisation, we create a set of heat maps showing pressure exerted on the cylinder test objects during a grip.  The cylinder was chosen due to its ability to be sensorised using UNEO GHF10-500N flexible pressure sensors (Fig.~\ref{fig:pressuresetup}).  The cylinder's circular profile matches the inner profile of the gripper,  allowing us to generate meaningful heatmaps (the cube, for example, could only have sensors mounted on its faces, yet typically sees significant forces exerted on the corners, where they would not be detected).
 
\begin{figure}[t!]
    \centering
    \includegraphics[width=1\linewidth]{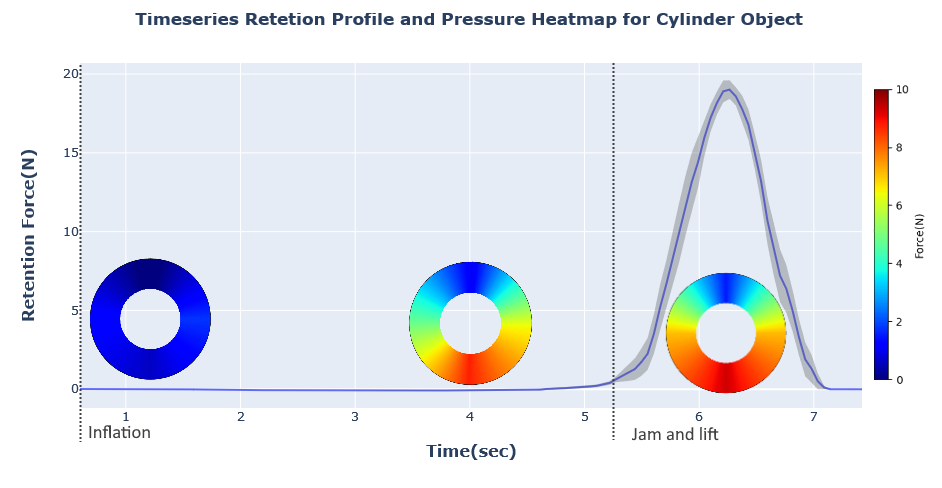}  
    \caption{Evolution of pressure when gripping a 30mm cylinder, as measured by 4 flexible pressure sensors.  The dark blue area at the top of the torus represents the joining part of the gripper, which aligns to the rear of the test rig.  The shaded region during a grip denotes standard error.}
    \label{fig:heatmaps}
\end{figure}

Four sensors are taped equidistantly around the cylinder. The gripper is oriented such that the joining component aligns with the rear of the testing rig and the rear sensor on the test object (the dark blue zones at the top of the torus in Fig.~\ref{fig:heatmaps}). The sensor readings are linearly interpolated to create the final plots. The test was repeated 10 times and the results averaged.

This test provides insights into the pressure evolution as the object is gripped. Pre-inflation we note minimal force exerted on the test object (leftmost heatmap, Fig~\ref{fig:heatmaps}.  Inflation exerts uneven pressure on the cylinder (centre heatmap): the joining segment has a greater wall thickness and the pressure inlets are resistant to deformation, so less surface contact is observed.  This is compensated by a concentration of pressure on the opposite side of the object (orange/red colour, peaking at 7.8N).  Jamming then occurs from 5s, reaching a mean grip force of 19.7N for an exerted force of 8.7N (rightmost heatmap).  The pressure is seen to spread more uniformly around the test object when jammed, as the jammed surface is forced against the walls of the object by ongoing inflation.  This creates a symmetric profile, indicating that the force is well distributed.

\begin{figure}[t!]
    \centering
    \includegraphics[width=1\linewidth]{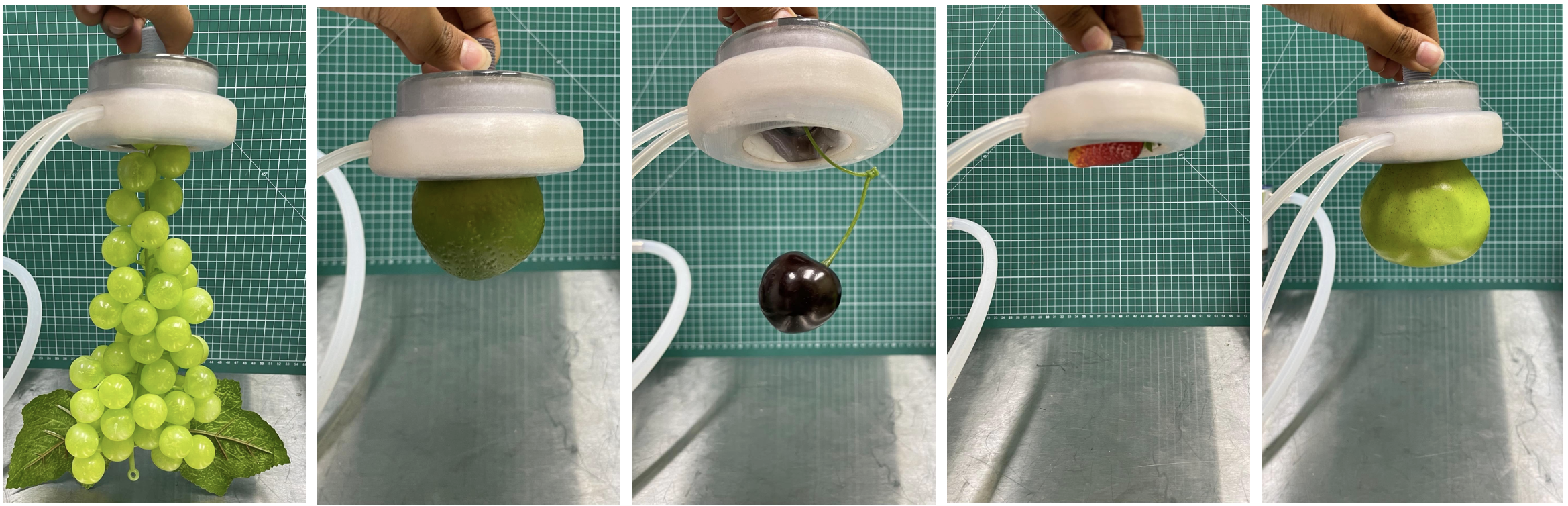}  
    \caption{Showing examples of fruit being grasped by the donut gripper.}
    \label{fig:fruity}
\end{figure}

Finally, we demonstrate some potential applications of the real-world picking ability of the gripper for soft produce (Fig.~\ref{fig:fruity}), including cherries and strawberries, and objects larger than the inner diameter of the gripper, including a bunch of grapes, limes, and pears.  All of these grips were created using the same control strategy and timing seen in Fig.~\ref{fig:heatmaps}, highlighting how even simple strategies can create workable grips across a range of different shapes and sizes of target object.
 
%% file: Content/6.0_Discussion.tex
\section{Conclusion}
The shift to soft robotics is evident within agriculture; however, the  actuation strength, lack of versatility and inadequate range of operation are open challenges for harvesting grippers.  We demonstrate the benefits of a hybrid free space soft gripper combining jamming and pneumatics.  Results demonstrate the ability of our gripper to generate significant grip strength across a range of different object sizes and shapes, which is key for agricultural applications.  
Crucially, our gripper provides the benefits of stiffness variation to achieve high retention force, without needing the target object to be sandwiched against a surface -- in other words our gripper does not require exertion of an activation force. This is beneficial for harvesting crops which are often hanging on branches within a lateral space without a surface to push against.

Further experimentation showed the benefits of our design compared to a traditional jamming gripper, and using different modes of operation as baselines for our gripper, showed the benefits of pushing a jamming component against the test object.  We also showed real world applications of picking various fruit with vastly different morphologies.  Pressure testing showed the ability of our gripper to spread grip strength across a wide surface area, reducing the risk of bruising soft produce at grip points.

Although this design focuses on smaller fruits like strawberries and cherries, the gripper is shown to generate solid grips on limes, pears, bunches of grapes, and other objects that are too large to be fully enveloped by the gripper.  This potentially opens up novel gripping strategies where a grip occurs only on specific parts of the target object -- this is the subject of future work.

\bibliographystyle{IEEEtran}
\bibliography{main}